\pdfoutput=1

\documentclass[11pt]{article}

\usepackage{ACL2023}

\usepackage{times}
\usepackage{latexsym}

\usepackage[T1]{fontenc}

\usepackage[utf8]{inputenc}

\usepackage{microtype}

\usepackage{inconsolata}

\usepackage{enumitem}
\usepackage{graphicx}
\usepackage{graphics}
\usepackage{url}
\usepackage{listings}
\usepackage{xspace}
\usepackage{amsmath}
\usepackage{booktabs}
\usepackage{caption}
\usepackage{subcaption}
\usepackage{amssymb}
\usepackage{pifont}
\usepackage{multirow}
\usepackage{tabularx}
\usepackage{longtable}
\usepackage{ltxtable}
\usepackage{placeins}
\usepackage[hang,flushmargin]{footmisc} 

\newcommand{\parheader}[1]{{\bf \smallskip \noindent #1.}}

\newcommand{\RankLLaMA}{RankLLaMA\xspace}
\newcommand{\RankGPT}{RankGPT\xspace}
\newcommand{\RankVicuna}{RankVicuna\xspace}
\newcommand{\LRL}{LRL\xspace}
\newcommand{\codellama}{Code-LLaMA-Instruct\xspace}
\newcommand{\ourmethod}{{Rank-wo-GPT}\xspace}
\newcommand{\msmarco}{MS~MARCO\xspace}

\newcommand{\ralph}[1]{\textcolor{cyan}{(#1)}}

\newcommand{\ignore}[1]{}

\newcommand{\QLoRA}{QLoRA\xspace}
\newcommand{\corank}{\texttt{co.rerank}\xspace}

\newcommand{\cross}[0]{\textcolor[HTML]{DA3F32}{\bf \ding{55}}\xspace}  
\newcommand{\tick}[0]{\textcolor[HTML]{398D64}{\bf \ding{51}}\xspace}  

\title{Rank-without-GPT: Building GPT-Independent Listwise Rerankers on Open-Source Large Language Models}

\author{Xinyu Zhang,$^{1}$\thanks{~~Work is done during internship at Cohere.}~~Sebastian Hofstätter,$^{2}$ Patrick Lewis,$^{2}$ Raphael Tang,$^{3}$ Jimmy Lin$^{1}$ \\[1ex]
$^1$University of Waterloo~~~$^2$Cohere~~~$^3$Comcast Applied AI\\
{\small $^1$\texttt{\{xinyucrystina.zhang, jimmylin\}@uwaterloo.ca}~~$^2$\texttt{\{sebastian, patrick\}@cohere.com}~~ $^3$\texttt{{raphael\_tang}@comcast.com}}
}

\begin{document}
\maketitle
\begin{abstract}
\textit{Listwise} rerankers based on large language models (LLM) are the zero-shot state-of-the-art.
However, current works in this direction all depend on the GPT models, 
making it a single point of failure in scientific reproducibility.
Moreover, it raises the concern that the current research findings only hold for GPT models but not LLM in general.
In this work, we lift this pre-condition and build for the first time effective listwise rerankers without any form of dependency on GPT.
Our passage retrieval experiments show that our best listwise reranker surpasses the listwise rerankers based on GPT-3.5 by 13\% and achieves 97\% effectiveness of the ones based on GPT-4.
Our results also show that
the existing training datasets, which were expressly constructed for \textit{pointwise} ranking, are insufficient for building such listwise rerankers. Instead, high-quality listwise ranking data is required and crucial,
calling for further work on building human-annotated listwise data resources.
\end{abstract}

\section{Introduction}
Given a user query,
the objective of \textit{text retrieval} is to fetch a list of documents among potentially billions of documents, ranked in descending order of relevance to the query. 
Mainstream solutions to this problem follow a multi-stage ranking pipeline.\
In the first stage, \textit{retrievers} are designed to efficiently retrieve the top-$k$ candidates from the entire collection,
followed in further stages by the application of \emph{rerankers}, which refine the ranking of the returned top-$k$ candidates.

Rerankers are traditionally constructed in a \textit{pointwise} paradigm, 
where given a query, the rerankers produce a relevance score for each passage independently,
and the final ranking is formed by sorting passages by their relevance scores.
Recently, 
attracted by the strong generative power of large language models (LLM) and their capacity to consume long-context inputs,
a new paradigm of neural rerankers has been proposed using \textit{listwise} ranking~\cite{lrl, rankGPT, rankVicuna, psc}.
These models consume a combined list of passages at a time
and directly outputs the reordered ranking list.\footnote{
Note that this is different from the listwise loss~\cite{listwiseloss}.
See details in Section~\ref{sec:listwise}.}

Not only does it achieve the state of the art on two TREC DL datasets~\cite{psc},
listwise ranking provides a novel perspective to passage reranking:\
this new paradigm questions the necessity to convert the ranking task into a classification task,
and instead frames it as a pure text generation task that could be solved end-to-end in a generalized text-to-text fashion~\cite{t5}.
For the first time,
the model directly generates the entire ranking list in the form of text,
instead of requiring multiple disjoint inference passes of the model as in pointwise~\cite{monobert, monoT5} or pairwise rerankers~\cite{prp, duot5}.
This integrates passage retrieval into the unified framework established in NLP,
and thus enables it to {merge} seamlessly with other text-to-text tasks and 
leverage existent prompting techniques~\cite{cot, liu2023lost}.

However,
while existing work on listwise reranking demonstrates the promising application of this new ranking paradigm,
their success \textit{crucially depends} on GPT models,
either directly for the inference-time model~\cite{lrl, rankGPT}
or indirectly for the training-time teacher model~\cite{rankVicuna}.
Such exclusive dependence results in a single point of failure in scientific reproducibility.
Moreover, it raises the concern that the current research findings are only applicable to the GPT models instead of the general LLMs.
In this work, we seek to reduce the reliance of listwise rerankers on GPT models and diversify the solution options for constructing such models.
Results show that, for the first time, our best listwise reranker built without any form of GPT dependence surpasses the rerankers based on GPT-3.5 by 13\% and achieves 97\% effectiveness of ones based on GPT-4,
measured by nDCG@10 on two passage retrieval datasets.

In this process, we found the current IR training data, which was constructed in order to train pointwise rerankers, is far from sufficient for training listwise rerankers~(\autoref{fig:dataformat}, Section~\ref{sec:pointwise-data-issue}),
yielding worse results than using data generated by BM25, a non-neural lexical technique in IR.
While silver ranking data generated by current rerankers serves as a good approximation of the gold ranking,
the performance of listwise rerankers increases linearly with training data ranking quality --- a relationship which has not yet plateaued (Section~\ref{sec:results}).
This indicates that the models are likely to further benefit from training data of higher quality,
calling for future work on building human-annotated datasets purpose-designed for listwise training.

The main purpose of our work is to advocate diverse solutions for future listwise reranking research.
Our contributions are as follows:\ 
\textbf{(1)}~We are first to show that the listwise rerankers, without any form of dependency on the GPT models, 
could outperform the listwise rerankers based on GPT-3~or~3.5 and perform on par with the ones based on GPT-4; 
\textbf{(2)}~We found that the ranking quality in the training data is crucial in constructing efficient listwise rerankers,
which might be the bottleneck of the current capacity of the listwise rerankers;
\textbf{(3)}~We demonstrate that listwise reranker fine-tuning is not data-expensive,
where an effective listwise reranker can be built using 5k queries, each associated with a list of passages ranked in high quality,
showing that it is feasible to build a human-annotated listwise dataset for this purpose.

\section{Background}
\ignore{
Prior to the era of large language models (LLMs), rerankers were finetuned in the \textit{pointwise} fashion,
where models receive a query and a single passage at a time and estimate the relevance score of the passages to the given query:
\vspace{-1mm}

\begin{small}
\begin{verbatim}
Input: [CLS] {query} [SEP] {passage} [SEP]
Output: 0.35
\end{verbatim}
\end{small}
\vspace{-1.5mm}

\noindent
Recently, attracted by the strong generative power of LLMs and their capacity to consume long-context inputs,
\textit{listwise} ranking is proposed, which receives a query and a list of passages at a time, and directly generates a reordered list of the input passages:

\vspace{-1mm}
\begin{small}
\begin{verbatim}
Input: {query} [1] {passage 1}, ..., [n] {passage n} 
Output: [5] > [3] > [4] > [2] > [1]
\end{verbatim}
\end{small}

This section introduces the details of the two methods, and then illustrates the issue of using current pointwise data in listwise fine-tuning. 
}

\subsection{Pointwise Reranking}
Given a query $q$ and a passage $p_i$, the pointwise reranker $h_\text{pw}$ produces a real score $s_i := h_\text{pw}(q, p_i)$ indicating the relevance of the passage to the query.
The model is optimized using cross entropy~\cite{monobert, monoT5}
or the contrastive loss~\cite{gao2021rethink, pradeep2022squeezing, rankT5, ma2023repllama},
based on binary relevance judgments from human annotators.

At inference time, given the top-$k$ passages $\{p_i\}_{i=1}^k$ returned by the previous-stage retriever, the model computes the relevance scores $\{s_i\}_{i=1}^k$ for each $p_i$ independently.
The final passages are then ranked by decreasing the magnitude of their corresponding relevance scores.

\begin{figure*}[t]
    \centering
    \includegraphics[width=1.45\columnwidth, trim={0mm 1.5mm 0mm 1.5mm},clip]{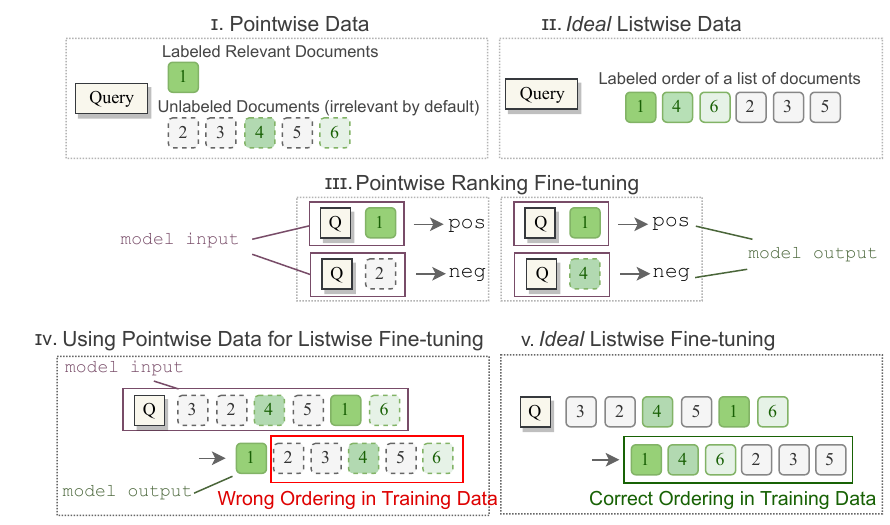}
    \caption{
        The issue with using current pointwise ranking data in listwise training. Numbers in the boxes indicate different passages.
        The grey boxes indicate irrelevant passages and the green ones indicate relevant ones.
        The saturation level indicates the relevance: the more saturating the green is, the more relevant the passages are.
        Boxes with dash borders indicate unlabeled passages, which are considered irrelevant in the current convention.
        Thus, the green boxes with dash borders are the false negative passages.
    }
    \label{fig:dataformat}
\end{figure*}

\subsection{Listwise Reranking}
\label{sec:listwise}
As opposed to pointwise rerankers, which rank passages according to their individual predicted relevance scores to the query,
listwise rerankers are designed to directly predict the final \textit{ranking} of a list of passages as a whole,
This not only allows the models to inter-reference the candidate passages to better determine their order, 
but also frames the passage retrieval task as text generation and thus {fuse} well with the existent techniques based on generative models. 
Using an LLM as a listwise reranker is concurrently studied in \RankGPT~\cite{rankGPT} and \LRL~\cite{lrl},
where both works use GPT-based models.

We formulate listwise rerankers under the same preliminaries as the pointwise one:\
given the instruction prompt $s$, the query $q$, and an input sequence of top-$k$ passages $\{p_i\}_{i=1}^k$, the listwise-ranking LLM $h_\text{lw}$ returns the final ranked passages $\hat{\mathcal{P}} := h_\text{lw}(q, \{p_i\}_{i=1}^k; s)$, where $\hat{\mathcal{P}}$ is a permutation (reranking) of $\{p_i\}_{i=1}^k$.

\parheader{Sliding window}
Limited by the maximum input length, we can feed only 10--20 passages to the LLM at a time. 
To rerank a longer list, e.g.\ typically top-100 passages,
both RankGPT and LRL adopt a \textit{sliding window strategy}, where we slide a window of size $n$ from the end to the front of the list and rerank the documents in the window, striding by $m$ documents per step.
In each stride, the top-$(n-m)$ documents are preserved and form the next sliding window, together with the next $m$ documents. 

\parheader{Fine-tuning listwise-ranking LLMs}
Used directly out of the box, current open-source LLMs often generate ill-formed outputs from listwise prompts~\cite{prp, rankVicuna},
where few valid ranking results can be inferred.
Thus, our work focuses on the condition of fine-tuning LLMs, which helps the models follow the instructions and generate valid outputs.
However, we found that the current human-annotated training data for IR is insufficient for this purpose, 
which we elaborate in Section~\ref{sec:pointwise-data-issue}.

\parheader{Difference from listwise loss}
Note that the listwise ranking mentioned in this work is different from the listwise loss in information retrieval (IR; ~\citealp{listwiseloss}),
where models still generate the score for each passage independently,
although the loss is computed by leveraging scores of a list of documents.
The term listwise in this work refers to that the model is capable of processing a list of documents at the same time.

\section{Method}
\subsection{Training Data for Listwise Reranker}
\label{sec:pointwise-data-issue}

The difference in the output format of the two above rerankers by nature requires different types of training data.
Past experience shows that a large-scale professionally annotated dataset with binary judgments, e.g., MS~MARCO~\cite{bajaj2016ms}, is sufficient in fine-tuning pointwise rerankers.
These pointwise datasets consist of queries, documents, and binary query--document labels, annotated to denote document relevance to the query.
Unannotated documents are considered irrelevant by default. (\autoref{fig:dataformat} Block~I, Block~III)

\ignore{ 
Intuitively, when using such data in listwise fine-tuning,
the list is formed by placing the labeled relevant documents in the front, which are then followed by the irrelevant ones.
}
However, there are challenges in constructing gold rankings using current resources for two main reasons.
First, there are many false-negative passages.
Taking \msmarco as an example, which is the largest training data in text retrieval, there is on average only one labeled passage per query.
In a list of, say, twenty retrieved passages, only one at most is known to be in the correct position (the first one), whereas the positions of the other nineteen are unknown.
This may result in an extremely noisy ordering.
Second, true relevance is nuanced and \textit{graded} (multilevel) rather than binary, as TREC evaluation sets show.
Binary relevance ignores nuances in the true relevance levels and discards the correct order of relevant passages, thus resulting in a suboptimal ordering of the passage list.
We concisely illustrate these two issues in \autoref{fig:dataformat} Block~IV.

\ignore{
One major source of this suboptimum comes from the false negative documents:
since only a few documents are labeled, there are likely to be a considerable number of relevant documents mixed in the unlabeled documents set.
Without correctly placing these documents in front of the truly irrelevant documents in the training set, 
the fine-tuning objectives would be noisy for LLM to learn.
(\autoref{fig:dataformat} Block~V)
\ralph{ 
Regardless, as a sanity baseline, we construct a list by placing the labeled relevant documents in the front and irrelevant irrelevant ones in the back.
}
}

\smallskip
To verify the above hypothesis that the ordering of the ranking list is crucial for fine-tuning listwise rerankers, 
we designed two sets of experiments:

\noindent
~~1. \textbf{Pointwise ground truth (P-GT)}:
    We construct a list by placing the labeled relevant documents in the front, which are then followed by the irrelevant ones ordered arbitrarily.
    This is used as a sanity baseline, showing the effectiveness when only using the human-annotated training data in the pointwise ranking manner.

\noindent
~~2. \textbf{Silver ranking}: we use the ranking results of several existent ranking systems to approximate the gold ranking. Specifically, we select the following ranking systems:\
\begin{enumerate}[leftmargin=0.75cm,topsep=0pt]
\itemsep0em 
    \item[a)] {\bf BM25}:
    Passages are ranked by BM25~\cite{robertson2009probabilistic}, a traditional unsupervised retrieval algorithm based on lexical matching.
    \item[b)] {\bf Fine-tuned Contriever (Contriever+ft)}:
    Passages are ranked by Contriever~\cite{contriever} that has been further fine-tuned on \msmarco. We used the checkpoint released by the original work.\footnote{\url{https://huggingface.co/facebook/contriever-msmarco}}
    \item[c)] {\bf \corank}: Passages are ranked by the Cohere rerank API.\footnote{\url{https://cohere.com/rerank}} We used the model \texttt{rerank-english-v2.0}.
\end{enumerate}



\noindent
The ranking systems are selected with increasing ranking capacity,
and thus generating listwise training data with increasing ranking quality.

\subsection{Prompt}
We adopt the same prompt as \RankGPT and \RankVicuna for a fair comparison of the results:\
\vspace{1em}

\noindent \textit{Input Prompt Template:}
\begin{small}
\begin{verbatim}
USER: I will provide you with {num} passages, each
indicated by a numerical identifier []. Rank the
passages based on their relevance to the search
query: {query}.
[1] {title 1} {passage 1}
[2] {title 2} {passage 2}
...
[{num}] {passage {num}}
Search Query: {query}.
Rank the {num} passages above based on their
relevance to the search query. All the passages
should be included and listed using identifiers, in
descending order of relevance. The output format
should be [] > [], e.g., [4] > [2]. Only respond
with the ranking results, do not say any word
or explain.
\end{verbatim}
\end{small}

\noindent \textit{Example Completion:}
\begin{small}
\begin{verbatim}
[4] > [5] > [2] > [3] > [1]
\end{verbatim}
\end{small}

\section{Experimental Setup}
\subsection{Models}
Most of the experiments in the work are conducted on \codellama~\cite{codellama},\footnote{\url{https://huggingface.co/codellama}}
given its transparency on model weights.
We experiment with all released model sizes: 7B, 13B, and 34B.
In ablation studies, we compare the results to {Vicuna-v1.5},\footnote{\url{https://huggingface.co/lmsys/vicuna-7b-v1.5}}
another model based on Llama 2, but then fine-tuned on ShareGPT, instructional data generated by GPT.

\begin{figure}
    \begin{subfigure}[t]{\columnwidth}
        \centering
        \caption{TREC-DL-19}
        \includegraphics[width=\textwidth, trim={0mm 2mm 0mm 5mm},clip]{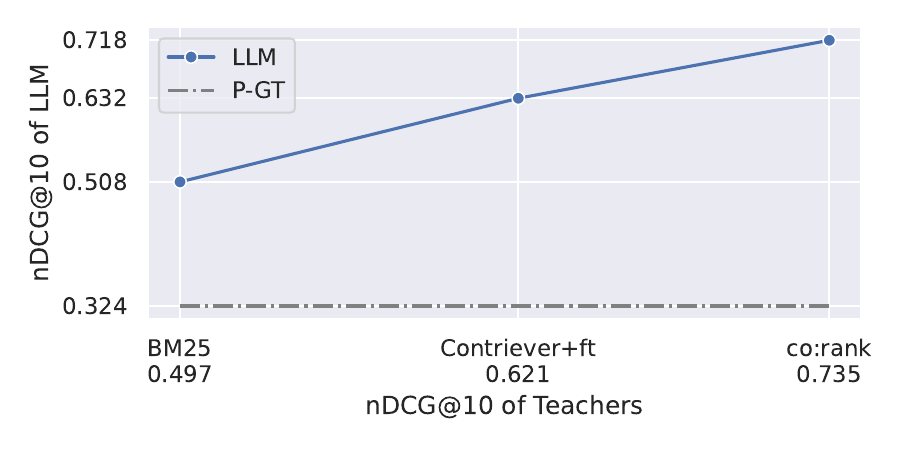}
    \end{subfigure}
   
    \begin{subfigure}[t]{\columnwidth}
        \centering
        \caption{TREC-DL-20}
        \includegraphics[width=\textwidth, trim={0mm 2mm 0mm 5mm},clip]{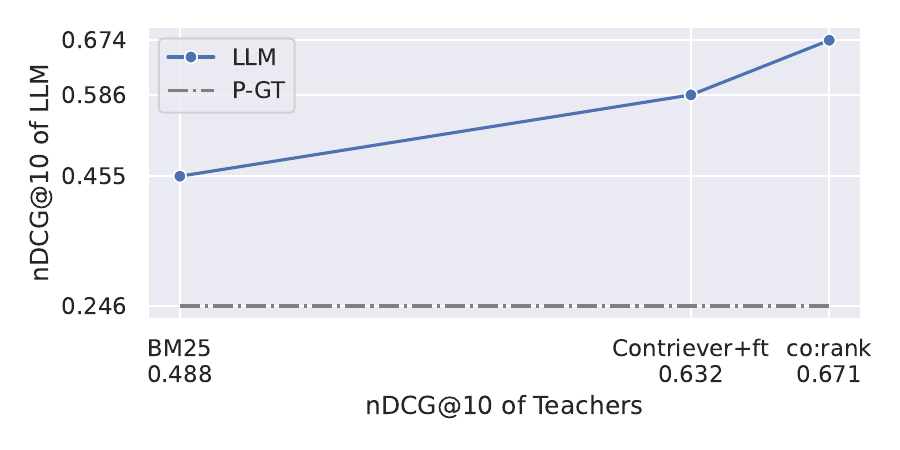}
    \end{subfigure}
    \caption{nDCG@10 on TREC-DL-19 and TREC-DL-20 when fine-tuned on data prepared on methods described in Section~\ref{sec:pointwise-data-issue}. 
    \textbf{P-GT}: Pointwise ground truth.
    }
    \label{fig:teacher-vs-student}
\end{figure}

\subsection{Data}
\label{sec:data}
\parheader{Training data preparation}
The training data are prepared from \msmarco v1 corpus~\cite{bajaj2016ms}, which contains 8.8 million passages.
We sampled $n$ training queries from the $100k$ training data of \RankVicuna ($n \in \{2k, 5k, 10k, 20k\}$),
then reordered the list of documents per query in the four settings mentioned in Section~\ref{sec:pointwise-data-issue}.

\parheader{Evaluation datasets}
We select TREC-DL-19 and TREC-DL-20~(\citealp{craswell2020overview}, \citeyear{craswell2021overview}) to evaluate the in-domain effectiveness. 
Both datasets are built from the TREC Deep Learning Track and share the same corpus with \msmarco v1~\cite{bajaj2016ms}.
In Section~\ref{sec:results:compare}, we report results reranking top-100 candidates returned by BM25~\cite{robertson2009probabilistic} and RepLLaMA~\cite{ma2023repllama}.

We report scores of nDCG@10 following the dataset standard.
In Section~\ref{sec:results:compare},
we also report some results of Judged@10,
the ratio of judged passages in the top-10 of the ranking list.

\begin{figure}
    \centering
    \begin{subfigure}[t]{\columnwidth}
        \caption{TREC-DL-19}
        \includegraphics[width=0.93\textwidth, trim={0mm 5mm 0mm 9mm},clip]{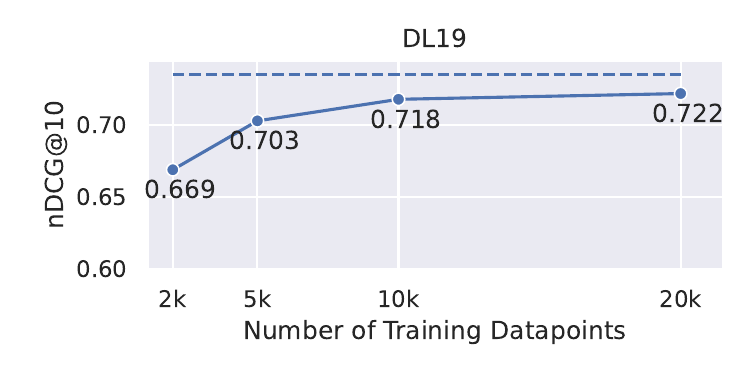}
    \end{subfigure}
   
    \begin{subfigure}[t]{\columnwidth}
        \caption{TREC-DL-20}
        \includegraphics[width=0.93\textwidth, trim={0mm 5mm 0mm 9mm},clip]{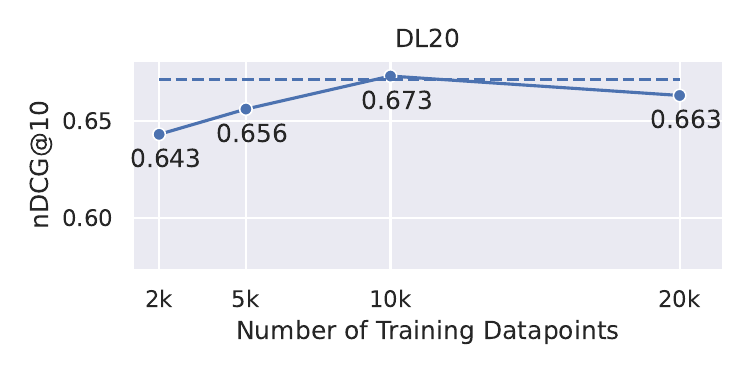}
    \end{subfigure}
    \caption{Results regarding the increasing number of training data generated by \corank. Dash lines refer to the result of \corank.}
    \label{fig:training-data-size}
\end{figure}

\subsection{Configurations}
In this work, we use FastChat~\cite{fastchat}\footnote{\url{https://github.com/lm-sys/FastChat}}
for the model training and inference.
FlashAttention~\cite{flashatten, flashatten2} is applied to all experiments. 
We turned on gradient checkpointing when fine-tuning 34B models.  
When not specified, we fine-tune the model with batch size 128.
The maximum input length is set as 4,096.
The fine-tuning epochs depend on the number of training datapoints.
The model is fine-tuned for 4 epochs when using 20k training data, 
8 epochs when using 10k training data, so on and on.

In experiments using QLoRA,
we set LoRA rank as 64, alpha as 16, dropout rate as 0.05, maximum gradient norm as 0.3,
and a constant learning rate of 1e-4, all following the advice from \citet{qlora}.
LoRA is applied on \texttt{q\_proj} and \texttt{v\_proj} layers.
In experiments that fine-tune the entire LLM,
we use a learning rate of $2\times10^{-5}$ with the cosine learning schedule. 
All experiments are run on 8 NVIDIA A100 GPUs with 80GB memory.
With \QLoRA, training 7B models takes around 5h when fine-tuning 20k training data for 4 epochs.

\section{Results and Analysis}
\label{sec:results}

\subsection{Training Data Quality}
\label{sec:results:quality}
We first show that the current pointwise labeled data alone could not serve the need of fine-tuning generative LLM as listwise rerankers. 
While the ranking results produced by current rerankers could be used as an approximation of the gold ranking,
the listwise rerankers are likely to further benefit from human-labeled listwise data in higher quality.

\autoref{fig:teacher-vs-student} shows the results on TREC-DL-19 and TREC-DL-20 of the listwise rerankers when fine-tuned on different training data.
The x-axis is the nDCG@10 of the pointwise rerankers that generate the training data,
and the y-axis is the nDCG@10 of the listwise rerankers fine-tuned on the corresponding data.
The horizontal dash line is the result when the model is fine-tuned on the ground-truth pointwise data only. 

Clearly, listwise rerankers fine-tuned only the pointwise data yield inferior ranking quality, evidenced by that the grey line is greatly lower than others.
When fine-tuned on the silver ranking data, the scores of the listwise rerankers follow closely to the scores of pointwise rerankers (e.g., scores on pointwise vs.\ corresponding listwise reranker: 0.497 vs.\ 0.508, 0.621 vs.\ 0.632, 0.735 vs.\ 0.718).
On one hand, this shows that the quality of rankings data is crucial when fine-tuning the listwise rerankers;
on the other hand, the listwise student is able to keep up with even one of the best current teachers without showing a trend of plateau.
This hints that the potential capacity of the listwise rankers may not be fully excavated and may be bounded by the quality of current training data.
That is, if higher-quality listwise training data were available (e.g., by human labeling), the listwise rankers may show higher ranking capacity.

\begin{figure}
    \begin{subfigure}[t]{\columnwidth}
        \centering
        \caption{TREC-DL-19}
        \includegraphics[width=0.93\textwidth, trim={0mm 5mm 0mm 9mm},clip]{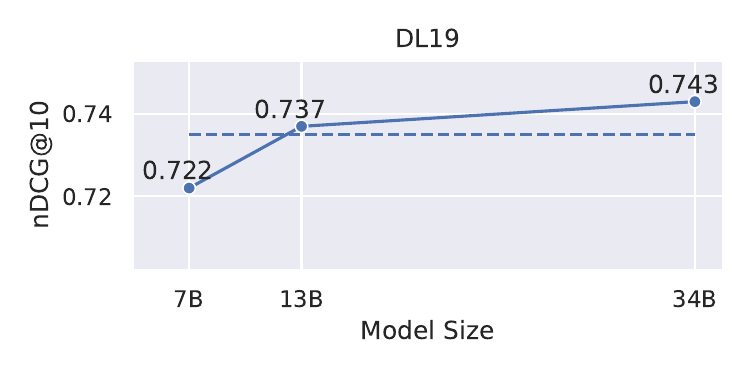}
    \end{subfigure}
   
    \begin{subfigure}[t]{\columnwidth}
        \centering
        \caption{TREC-DL-20}
        \includegraphics[width=0.93\textwidth, trim={0mm 5mm 0mm 9mm},clip]{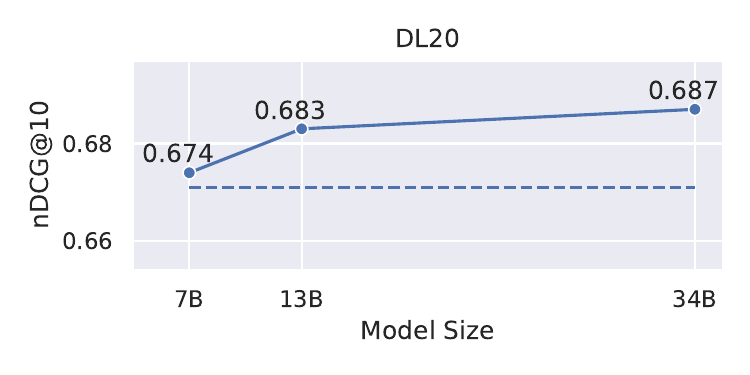}
    \end{subfigure}
    \caption{
        Result regarding different sizes of the model, all fine-tuned on 10k data.
        Dash lines refer to the result of \corank. 
    }
    \label{fig:model-size}
\end{figure}
\begin{table*}[t]
\resizebox{\linewidth}{!}{ \begin{tabular}{llcrlrcc}
\toprule
     & \multicolumn{1}{c}{\multirow{2}{*}{\textbf{Model}}} & \multicolumn{1}{c}{\textbf{GPT-}} & \multicolumn{1}{c}{\textbf{Model}} & \multicolumn{1}{c}{\textbf{Previous}} & \multicolumn{1}{c}{\multirow{2}{*}{\textbf{top-$k$}}}  & \multicolumn{1}{c}{\textbf{TREC-DL-19}} & \multicolumn{1}{c}{\textbf{TREC-DL-20}} \\

     & & \textbf{independent} & \multicolumn{1}{c}{\textbf{Size}} & \multicolumn{1}{c}{\textbf{Stage}} & & \multicolumn{1}{c}{nDCG@10} & \multicolumn{1}{c}{nDCG@10} \\ 
\midrule
    \multicolumn{8}{c}{\textit{non-listwise methods based on non-LLM}} \\
    (1)~~ monoBERT & BERT & \tick & 110M & BM25 & 1000 & 72.3 & 72.2 \\
    (2)~~ monoT5 & T5 & \tick & 3B & BM25 & 100 & 71.8 & 68.9 \\
    (3)~~ rankT5 & T5 & \tick & 3B & BM25 & 100 & 71.2 & 69.5 \\

\midrule
    \multicolumn{8}{c}{\textit{non-listwise methods based on LLM}} \\
    (4)~~ UPR & FLAN-T5-XXL & \tick & 11B & BM25 & 100 & 62.0 & 60.3 \\
    (5)~~ PRP-Sliding-10 & FLAN-UL2 & \tick & 20B & BM25 & 100 & 72.7 & 70.5 \\
    (6)~~ \RankLLaMA & LLaMA & \tick & 7B & RepLLaMA & 100 & 75.3 (\textit{76.1}) & 76.7~(\textit{76.2}) \\

\midrule
    \multicolumn{8}{c}{\textit{listwise methods}} \\
    (7)~~ RankVicuna & Vicuna & \cross & 7B & BM25 & 100 & 66.8 & 65.5 \\
    (8)~~ LRL & GPT-3 & \cross & ? & BM25 & 100 & 65.8 & 62.2 \\
    (9)~~ RankGPT-3.5 & GPT-3.5 & \cross & ? & BM25 & 100 & 65.8 & 62.9 \\
    (10)~RankGPT-4 & GPT-4 & \cross & ? & BM25 & 100 & 75.7 & 71.0 \\
\cmidrule{1-8}
    (11)~\textit{\ourmethod} & \multirow{4}{*}{\codellama} & \tick & 7B &  BM25 & 100 & 71.8~(\textit{70.8}) & 67.4~(\textit{66.7}) \\
    (12)~\textit{\ourmethod} & & \tick & 7B &  RepLLaMA & 100 & 73.0~(\textit{75.2}) & 70.0~(\textit{71.7}) \\
    (13)~\textit{\ourmethod} & & \tick & 13B & BM25 & 100 & 73.7 & 68.3 \\
    (14)~\textit{\ourmethod} & & \tick & 34B & BM25 & 100 & 74.3 & 68.7 \\

\bottomrule
\end{tabular}}
\caption{
Comparison of listwise reranker fine-tuned on data generated by \corank to other methods in the field,
evaluated on TREC-DL-19 and TREC-DL-20.
The \textit{tilted} scores in bracket are the ones evaluated on enriched query--passage relevance judgment, with Judged@10 $=1$.
Results of RankVicuna, LRL, and RankGPT-3.5 are copied from the original paper~\cite{rankVicuna, lrl, rankGPT}.
Results of RankGPT-4 reranking BM25 top-100 are copied from \citet{psc}.
}
\label{tab:main}
\end{table*}

\subsection{Training Data Quantity}
Having proved that higher-quality data is necessary to obtain effective listwise rerankers,
we ask the next question: \textit{how much data is required?}
\autoref{fig:training-data-size} compares the model effectiveness with increasing amounts of fine-tuning data.
For a fair comparison, the models are fine-tuned for the same number of steps when varying training data quantity:\
given that the model is fine-tuned for 8 epochs on 10k datapoints,
it is then fine-tuned for 40, 16, and 4 epochs when using 2k, 5k, and 20k datapoints,
where each datapoint consists of one query and 20 passages.
Therefore, training with fewer datapoints only saves the anticipated human labor effort for annotation but not the training time. 
Experiments are based on \codellama in size 7B.

As \autoref{fig:training-data-size} shows,
training on 5k training datapoints already yield 97\% of the effectiveness compared to using 10k datapoints,
whereas increasing data quantity from 10k to 20k only brings marginal improvement in the case of TREC-DL-19 and no positive effect on TREC-DL-20.
That is, 100k high-quality query--passage pairs (5k queries with 20 passages per query) serve the need of effectively fine-tuning listwise rerankers.
This is in the same scale with fine-tuning pointwise rerankers,
where \RankLLaMA~\cite{ma2023repllama} consumes 300k query--passage pairs from MS MARCO.

\subsection{Model Size}
The experiments above are all based on \codellama with size 7B.
We then examine the effect of scaling up the models.
As expected, the effectiveness of the listwise rerankers increases with the language model size.
\autoref{fig:model-size} shows the trend of the ranking quality with respect to the model size,
where the model of 13B already outperforms the teacher,
and increasing the model size to 34B brings additional improvement.

\subsection{Comparisons with Other Baselines}
\label{sec:results:compare}
Finally,
we compare our listwise rerankers to other methods in the field, evaluated on TREC-DL-19 and TREC-DL-20.
Results are shown in \autoref{tab:main}.
The baselines are grouped into three categories:\
\textbf{(1)} non-listwise rerankers based on non-LLM models (e.g., BERT);
\textbf{(2)} non-listwise rerankers based on LLM, including methods based on query likelihood~\cite{upr}, pairwise~\cite{prp} and pointwise reranking~\cite{ma2023repllama}; 
\textbf{(3)} listwise rerankers~\cite{rankVicuna, lrl, rankGPT, psc}, which all depend on GPT models.

\parheader{Unlabeled top-reranked passages}
Although TREC-DL data have comparatively dense human judgments,\footnote{120 judgments per query on TREC-DL-19; 211 judgments per query on TREC-DL-20}
we observe that listwise rerankers bring more unlabeled passages to the top of the reranked list compared to the pointwise ones.
For example, on TREC-DL-19,
the Judged@10 of listwise rerankers are between 0.88 to 0.94,
whereas the Judged@10 of \RankLLaMA is over 0.98.

For a fair comparison,
we manually annotated the missing query--passage relevance judgments from the top-10 of the lists returned by some of the rerankers, including both pointwise and listwise ones from rows~(6, 11, 12). 
The labels are on the same scale as the original graded judgment (i.e., from 0 to 3, with larger numbers indicating higher relevance). 
These added labels, together with the initial ones,
form the new judgment set,
which we refer to as ``\textit{enriched judgments}''.

Scores evaluated on our enriched judgments set are \textit{italicized} in parentheses.
We observe that the added judgment made a nontrivial difference to the evaluation results.
Most prominently,
the nDCG@10 on row~(12) increased from 73.0 to 75.2 after filling in the missing relevance.
Intact judgments also amend the over-rated rankings,
for example, on row~(11), 
the scores decreased with more labels. 
In the rest of this section,
we compare results evaluated on the enriched judgments.

\parheader{Comparison to GPT-based listwise rerankers}
Comparing rows~(11, 13, 14) to rows~(7--10), we found even our smallest listwise reranker (7B) is significantly higher than previous models based on GPT-3 and GPT-3.5.
Compared to RankGPT-4,
while the listwise rerankers yield lower scores with models of all sizes, 
the difference is again insignificant (two-tailed t-test, with $p < 0.01$).

\parheader{Comparison to LLM-based pointwise rerankers}
While the pointwise rerankers are fine-tuned on the optimal human-annotated data,
we find our listwise models,
fine-tuned under data non-optimized for its purpose,
perform close to the best pointwise rerankers in the same model size on TREC-DL-19.
Comparing row~(12) to row~(6), 
where both listwise and pointwise rerankers are based on the same size of models (7B) and reranking the same candidates from the first-stage retriever,
there is only a small gap between the nDCG@10 on TREC-DL-19,
with \textit{insignificant} difference (two-tailed t-test, with $p < 0.01$),
although there is a larger gap on TREC-DL-20: 71.7 vs.\ 76.2 on rows~(12, 6), with a significant difference. 
It would require future work to close the gap between the two.

\begin{figure}
    \centering
    \begin{subfigure}[t]{0.49\columnwidth}
        \centering
        \includegraphics[width=\textwidth, trim={255mm 0mm 5.5mm 8mm},clip]{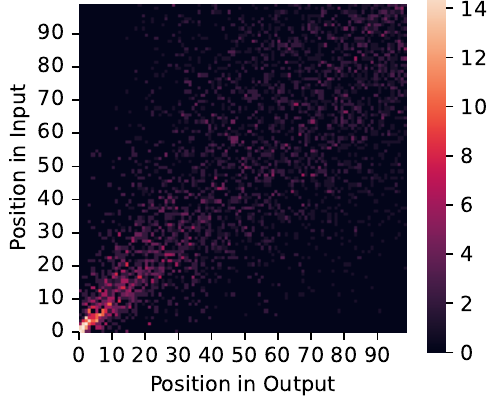}
        \caption{Pointwise Reranker}
        \label{fig:io-position:pointwise}
    \end{subfigure}
    \hfill%
    \begin{subfigure}[t]{0.49\columnwidth}
        \centering
        \includegraphics[width=\textwidth, trim={255mm 0mm 5mm 8mm},clip]{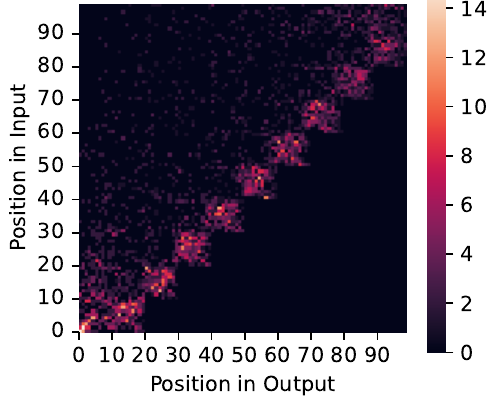}
        \caption{Listwise Reranker}
        \label{fig:io-position:listwise}
    \end{subfigure}
    \caption{
        Compare the position of relevant passages before and after reranking by RankLLaMA and \ourmethod
        both reranking RepLLaMA top-100.
        x-axis: passage positions in the reranked list;
        y-axis: passage positions in the first-stage list.
        Best viewed in color.
    }
    \label{fig:io-position}
\end{figure}

\subsection{Analysis on Sliding Window Strategy}
\label{sec:analysis}
While the sliding window strategy is a natural resort to apply listwise ranking on a passage list longer than the model input capacity,
it is unknown yet how well it aggregates the list in each pass. 

To start answering this question, we plot the ranking positions of relevant passages before and after reranking.
\autoref{fig:io-position} compares the position difference when using the pointwise and listwise rerankers,
the models on rows~(6) and (12) in \autoref{tab:main}.
In each heatmap, 
the y-axis indicates the passage position in the first-stage ranking (i.e., RepLLaMA) and the x-axis indicates the position after reranking by \RankLLaMA~(\autoref{fig:io-position:pointwise}) or \ourmethod~(\autoref{fig:io-position:listwise}).

Comparing the heatmaps,
we observe a prominent pattern in the listwise heatmap (\autoref{fig:io-position:listwise}) that there is a chain of bright clusters in the square shape along the diagonal line.
This indicates that a large number of relevant documents are ``trapped'' in the local block, promoted only within the current or the next pass of the sliding window.
This phenomenon is common for relevant passages at all relevant levels (Appendix~\ref{ap:analysis-full}, \autoref{fig:io-position-full}).

The brightness density in the upper matrix indicates the frequency of relevant passages promoted over a long distance over the list.
Compared to pointwise, where the scatters distribute symmetrically along the diagonal matrix,
listwise heatmap shows more scatters clustered in left-most columns, $x \in [0, 20]$,
indicating that the top-ranked passages by listwise rerankers still come from a wider range of positions in the first-stage results compared to the pointwise methods
regardless that a large number of passages are trapped as aforementioned.

\section{Ablation Studies}

\parheader{LLM with GPT-based instruction fine-tuning}
To investigate if more GPT-alike instruction fine-tuning would further benefit the listwise ranking results,
we ran the same experiment on Vicuna-v1.5.
As shown in rows~(1, 2) in \autoref{tab:qlora-vs-full},
while fine-tuning based on Vicuna achieved slightly better results on both datasets,
the difference is not significant.
Thus we conclude starting from a GPT-free LLM is able to yield satisfactory effectiveness compared to a more GPT-alike LLM.

\parheader{Fine-tuning Full Model vs.\ \QLoRA}
In previous experiments, we fine-tuned the LLM using \QLoRA instead of the entire LLM model to alleviate the GPU memory and disk requirement.
Here, we compared the effectiveness of the two fine-tuning strategies on Vicuna.\footnote{We conducted the same experiment in \codellama, however, the results were not in the correct scale. Thus we use Vicuna as a replacement in this ablation.}
As shown in rows~(2, 3) in \autoref{tab:qlora-vs-full}, 
fine-tuning with \QLoRA yields similar effectiveness as fine-tuning all parameters on both datasets, with the same amount of training data and the fine-tuning epochs.

\begin{table}[t]
    \centering
    \resizebox{\columnwidth}{!}{\begin{tabular}{rlc|cc}
        \toprule
               & & \textbf{Model} & \textbf{DL-19} & \textbf{DL-20} \\
        \midrule
         (1) & \QLoRA & \codellama & 0.718 & 0.674 \\
         (2) & \QLoRA & Vicuna-v1.5 & 0.728 & 0.683 \\
         (3) & Full & Vicuna-v1.5 & 0.727 & 0.674 \\
        \bottomrule
    \end{tabular}}
    \caption{
       Results when using \codellama and Vicuna as the initial LLM,
       and when fine-tuning Vicuna with \QLoRA or all parameters (Full).
       All models are in size 7B and fine-tuned on 10k datapoints for 8 epochs.
    }
    \label{tab:qlora-vs-full}
\end{table}

\section{Related Work}

In the past few years,
the question of how generative models could bring benefits to information retrieval has been an area of intense study, with a number of differing and complementary techniques emerging.
The strong generative performance of LLMs has been leveraged for retrieval by generating a large volume of synthetic datasets on domains: InPars~\cite{inpars, inparsV2}, and Promptagator~\cite{Promptagator}.

In parallel, researchers have investigated whether LLMs could be used directly as retrievers or rerankers: 
SGPT~\cite{sgpt} first shows that the GPT-based decoder models,
are effective when used as bi-encoder in retrieval tasks. 
UPR~\cite{upr} uses the query likelihood as the ranking score.
PRP~\cite{prp} shows that the LLM can effectively determine the comparative relevance regarding the query, given a pair of documents.
Recently, \citet{ma2023repllama}~demonstrate that fine-tuning LLAMA in the traditional paradigm of bi-encoder and pointwise cross-encoder surpasses smaller models.

Finally, a line of work that is mostly relevant to our work regards LLMs as black boxes and only uses the final generative output for ranking:\
\RankGPT~\cite{rankGPT} and \LRL~\cite{lrl} studied listwise rerankers concurrently, demonstrating its effectiveness using GPT-3, GPT-3.5, and GPT-4.
RankVicuna~\cite{rankVicuna} then showed that the method could be applied to a smaller-sized open-source LLM (e.g.\ Vicuna~\cite{vicuna} in 7B, 13B) by distilling from GPT-3.5.
\citet{psc} proposed a permutation self-consistency prompting method, which alleviates the positional bias and largely improves the effectiveness of the listwise ranking.

\section{Conclusions and Future Work}
In this work, we study how to construct effective \textit{GPT-free} listwise rerankers based on open-source LLM models.
Experiments on two passage retrieval datasets show that our listwise rerankers, without any form of dependency on GPT, can substantially outperform the ones built on GPT-3 and perform on par with the ones built on GPT-4.

In this process, we find that current pointwise training data in IR is not sufficient in fine-tuning listwise rerankers. 
Instead, training data comprised of high-quality ranked document lists is required and crucial.
While the training data generated by current pointwise rerankers could be used as a nice approximation,
the models are likely to benefit more from higher-quality listwise training data that are built from human annotations.

We hope this work sets up the stage for future research on the listwise ranking methods by bringing more diversity of the solutions to the research in this line.
Additionally, we hope it paves the path for future work on addressing text retrieval in the text generation paradigm,
where it could be formatted in the same way as the other text-to-text tasks, and thus better integrated into the unified system.

\section*{Limitations}
Despite the new paradigm brought by listwise ranking,
it still has intrinsic issues that are pending to be solved.
The most prominent one is the query latency:\
since the current listwise rerankers depend on the sequential inference of models with extremely large sizes,
they intrinsically suffer from higher query latency than rerankers based on BERT-size models.

\section*{Acknowledgement}
We thank Ronak Pradeep for providing the data of \RankVicuna,
Xueguang Ma for providing the runfiles of RepLLaMA and \RankLLaMA,
and Pat Verga and Jiarui Xu for their helpful discussion and advice on the paper.

\bibliography{custom}
\bibliographystyle{acl_natbib}

\appendix

\begin{table*}[hbt!]
\resizebox{\textwidth}{!}{ 
\begin{tabular}{lrrrrrrrrrr}
\toprule
& \textbf{BM25} & \textbf{GTR-XXL} & \textbf{cpt-text-XL} & \textbf{Ada2} & \textbf{SGPT} & \textbf{RepLLaMA} & \textbf{RankT5} & \textbf{RankLLaMA} & \textbf{RankLLaMA} & \textbf{\ourmethod} \\
\midrule
Model Size & -- & 4.8B & 175B & ? & 5.8G & 7B & 220M & 7B & 13B & 7B \\
\midrule
\textbf{DBPedia} & 31.8 & 40.8 & 43.2 & 40.2 & 39.9 & 43.7 & 44.2 & 48.3 & 48.7 & 42.3 \\
\textbf{FiQA} & 23.6 & 46.7 & 51.2 & 41.1 & 37.2 & 45.8 & 44.5 & 46.5 & 48.1 & 35.1 \\
\textbf{NF Corpus} & 32.2 & 34.2 & 40.7 & 35.8 & 36.2 & 37.8 & 38.1 & 30.3 & 28.4 & 32.8 \\
\textbf{SCIDOCS} & 14.9 & 16.1 & \multicolumn{1}{l}{-} & 18.6 & 19.7 & 18.1 & 18.1 & 17.8 & 19.0 & 16.2 \\
\textbf{SciFact} & 67.9 & 66.2 & 75.4 & 73.6 & 74.7 & 75.6 & 75.0 & 73.2 & 73.0 & 64.7 \\
\textbf{TREC-COVID} & 59.5 & 50.1 & 64.9 & 81.3 & 87.3 & 84.7 & 80.7 & 85.2 & 86.1 & 80.4 \\
\cmidrule{1-11}
\textbf{Average} & 38.3 & 42.4 & 55.1 & 48.4 & 49.2 & 51.0 & 50.1 & 50.2 & 50.6 & 45.2 \\
\bottomrule
\end{tabular}
}
\caption{nDCG@10 scores on BEIR subset}
\label{tab:beir}
\end{table*}
\begin{figure*}
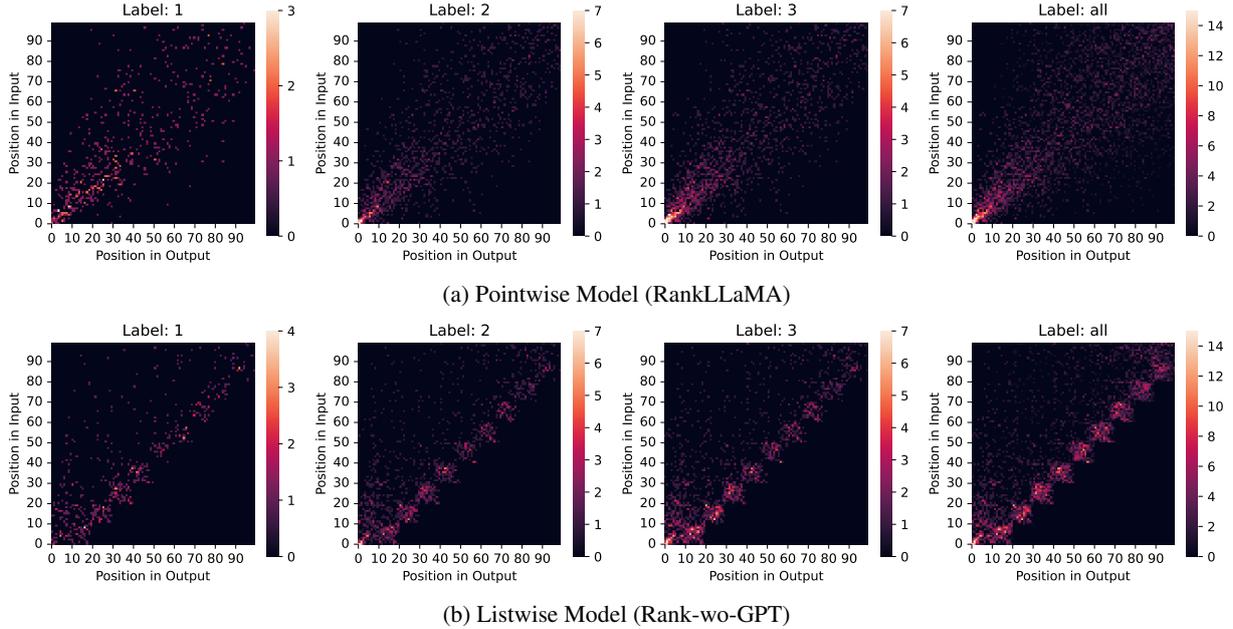

    \centering
    \begin{subfigure}[b]{\textwidth}
        \centering
        \includegraphics[width=\textwidth, trim={4mm 0mm 5mm 0mm},clip]{resources/rankllama-io-position.pdf}
        \caption{Pointwise Model (RankLLaMA)}
    \end{subfigure}

    \begin{subfigure}[b]{\textwidth}
        \centering
        \includegraphics[width=\textwidth, trim={4mm 0mm 5mm 0mm},clip]{resources/listwise-io-position.pdf}
        \caption{Listwise Model (\ourmethod)}
    \end{subfigure}
    \caption{
        The full version of \autoref{fig:io-position}, showing a separate heatmap for documents at each relevant level. 
        Labels from 1 to 3 indicate the passage has increasing relevance to the query.
        x-axis: the passage positions in the reranked list;
        y-axis: the passage positions in the first-stage list.
    }
    \label{fig:io-position-full}
\end{figure*}

\FloatBarrier
\section{Out-of-Domain Results}
In this section,
we share the out-of-domain results on BEIR~\cite{beir} of the 7B model, reranking the top-100 passages returned by BM25. 
Unfortunately, we only observe an unsatisfied generalization ability of the current model.
This requires future work on adapting fine-tuned listwise rerankers into different domains.
Results are reported in \autoref{tab:beir}.

\section{Heatmaps of Sliding Window Strategy on All Relevance Levels}
\label{ap:analysis-full}
In Section~\ref{sec:analysis}, we share the visualization of all the relevant passages positions before and after reranking. 
In \autoref{fig:io-position-full}, we show similar visualization separately for passages in different relevance level,
showing that the trend is similar overall.  

\section{Additional Annotation on TREC-DL}
As mentioned in Section~\ref{sec:results:compare},
the authors manually annotated the unlabeled query--passage pairs in top-10 returned by some rerankers. 
We share the annotation results in the following pages.
\onecolumn
\newcolumntype{Y}{>{\small\raggedright\arraybackslash}X}
\renewcommand{\arraystretch}{1.5}
\noindent

\small
\begin{longtable}{>{\setlength\hsize{.11\hsize}}X>{\setlength\hsize{1.89\hsize}}X}
\multicolumn{2}{l}{\textit{Query 1063750: why did the us volunterilay enter ww1}} \\
\toprule
\multicolumn{2}{c}{\textit{Label 0}} \\[-0.5em]
1126299 & World War 1 : How did WW1 affect the United States? The United States was affected by World War 1. The U.S was effected by political/social problems, racial conflicts and foreign policy issues. For example, the United States was effected by Isolationism, the Red Scare, and the Red Summer. The United States was affected by foreign policy issues with Isolationism. \\
\bottomrule
\multicolumn{2}{l}{} \\[-1.5em]

\multicolumn{2}{l}{\textit{Query 1114819: what is durable medical equipment consist of}} \\
\toprule
\multicolumn{2}{c}{\textit{Label 3}} \\[-0.5em]
1724525 & List of Durable Medical Equipment Items. This is a list of items you may need to use to manage your health or the health of a loved one. Many of these items are covered by Medicare. Air Cleaners. Air Conditioners. Air-Fluidized Bed. Alternating Pressure Pads and Mattresses. Audible/Visible Signal Pacemaker Monitor. Augmentative Communication Device. Bathtub Lifts. \\
\bottomrule
\multicolumn{2}{l}{} \\[-1.5em]

\multicolumn{2}{l}{\textit{Query 1115776: what is an aml surveillance analyst}} \\
\toprule
\multicolumn{2}{c}{\textit{Label 0}} \\[-0.5em]
5483591 & A professional designation awarded by the Association of Certified Anti-Money Laundering Specialists (ACAMS) to anti-money laundering professionals who have 40 qualifying credits based on education, other professional certification and work experience, who pass the CAMS examination and who provide three professional references. \\
\bottomrule
\multicolumn{2}{l}{} \\[-1.5em]

\multicolumn{2}{l}{\textit{Query 1129237: hydrogen is a liquid below what temperature}} \\
\toprule
\multicolumn{2}{c}{\textit{Label 3}} \\[-0.5em]
4254816 & Hydrogen is a liquid at -253 degree C and it is very near to absolute zero which is known as -273 degree C. Edit. Share to: 1  The Periodic Table of Elements Life is sustained by a number of chemical elements. 2  They are responsible for the birth of planets and the evolution of cells to complex organisms. \\
\bottomrule
\multicolumn{2}{l}{} \\[-1.5em]

\multicolumn{2}{l}{\textit{Query 130510: definition declaratory judgment}} \\
\toprule
\multicolumn{2}{c}{\textit{Label 0}} \\[-0.5em]
5668642 & declaration. n. 1. (Rhetoric) an explicit or emphatic statement. 2. (Rhetoric) a formal statement or announcement; proclamation. 3. (Rhetoric) the act of declaring. 4. (Law) the ruling of a judge or court on a question of law, esp in the chancery division of the High Court. \\
\bottomrule
\multicolumn{2}{l}{} \\[-1.5em]

\multicolumn{2}{l}{\textit{Query 182539: example of monotonic function}} \\
\toprule
\multicolumn{2}{c}{\textit{Label 0}} \\[-0.5em]
1787567 & Example: the function v(x) = x3 - x2 + 4x. To move C spaces to the left, add C to x wherever x appears: w(x) = (x + C)3 - (x + C)2 + 4(x + C) An easy way to remember what happens to the graph when we add a constant: add to y to go high. add to x to go left. \\
\bottomrule
\multicolumn{2}{l}{} \\[-1.5em]

\multicolumn{2}{l}{\textit{Query 405717: is cdg airport in main paris}} \\
\toprule
\multicolumn{2}{c}{\textit{Label 1}} \\[-0.5em]
7288990 & Charles de Gaulle Airport is located within portions of several communes 25 km (16 mi) to the northeast of Paris. Charles de Gaulle Airport serves as the principal hub for Air France as well as a focus city for low-cost carriers Vueling and Norwegian Air Shuttle. \\
\bottomrule
\multicolumn{2}{l}{} \\[-1.5em]

\multicolumn{2}{l}{\textit{Query 87452: causes of military suicide}} \\
\toprule
\multicolumn{2}{c}{\textit{Label 1}} \\[-0.5em]
8333913 & But the military will have to change the way it handles soldiers with mental illness if we expect to see the number of suicides decline, according to the author of a related editorial published in the journal. VA boss: 'No veteran should have to wait for claims'. \\
\bottomrule
\multicolumn{2}{l}{} \\[-1.5em]

\multicolumn{2}{l}{\textit{Query 1106007: define visceral?}} \\
\toprule
\multicolumn{2}{c}{\textit{Label 3}} \\[-0.5em]
746358 & Visceral, which means “pertaining to the soft organs in the abdomen,” is the fat stored deep in our abdomens around the intestines, kidneys, pancreas and liver. This is the stuff that tends to make our tummies protrude in classic “beer belly” fashion. \\
\multicolumn{2}{c}{\textit{Label 0}} \\[-0.5em]
2186171 & Visceral (Smooth) Muscle. Visceral muscle is found in various parts of the body including blood vessels, the bladder, digestive tract, as well as in many other hollow organs. Like cardiac muscle, most visceral muscle is regulated by the autonomic nervous system and is under involuntary control. \\
\bottomrule
\multicolumn{2}{l}{} \\[-1.5em]

\multicolumn{2}{l}{\textit{Query 87181: causes of left ventricular hypertrophy}} \\
\toprule
\multicolumn{2}{c}{\textit{Label 3}} \\[-0.5em]
3363613 & An enlarged left ventricle is usually due to either cardiomyopathy (a weakening of the heart muscle) or valvular heart disease. Valve disease is usually treated by correcting the underlying valve problem (initially with medicines, often later with surgery). The most common cause of an enlarged left ventricle is cardiomyopathy. Initial treatment is with medications, such as diuretics, digitalis, vasodilators (ACE inhibitors and/or ARB inhibitors), and beta blockers, such as carvedilol (Coreg) or metoprolol (Lopressor, Toprol XL). \\
\multicolumn{2}{c}{\textit{Label 0}} \\[-0.5em]
3216319 & However, high blood pressure, or hypertension, is the most common cause by far. Higher pressures within blood vessels put a strain on the ventricle, making it difficult to satisfy the body's needs. Also, narrowing of the aortic valve, or stenosis, obstructs blood flow from the left ventricle. In patients with HCM, the ventricle can become thick and stiff, which adds extra strain on the heart. \\
\bottomrule
\multicolumn{2}{l}{} \\[-1.5em]

\multicolumn{2}{l}{\textit{Query 527433: types of dysarthria from cerebral palsy}} \\
\toprule
\multicolumn{2}{c}{\textit{Label 0}} \\[-0.5em]
2722036 & Cerebral palsy is a group of disorders that can involve brain and nervous system functions, such as movement, learning, hearing, seeing, and thinking. There are several different types of cerebral palsy, including spastic, dyskinetic, ataxic, hypotonic, and mixed. \\
6015377 & Dysarthria is a motor speech disorder resulting from neurological injury of the motor component of the motor-speech system and is characterized by poor articulation of phonemes (cf. aphasia: a disorder of the content of language). \\
7748146 & Cerebral palsy is a general term describing conditions that cause movement problems. The most common type is spastic cerebral palsy where the muscles are stiff and rigid in one or more limbs. The underlying problem is damage to, or faulty development of, part of the brain. \\
\bottomrule
\multicolumn{2}{l}{} \\[-1.5em]

\multicolumn{2}{l}{\textit{Query 148538: difference between rn and bsn}} \\
\toprule
\multicolumn{2}{c}{\textit{Label 3}} \\[-0.5em]
1516299 & A registered nurse (RN) is a nurse who has passed the National Council Licensure Examination for Registered Nurses (NCLEX-RN) after completing an accredited training program. These programs come in the form of an Associate of Science in Nursing (ASN) or a Bachelor of Science in Nursing (BSN). The degrees differ in the time it takes to complete them and the type of coursework required for graduation. \\
332406 & RN vs. BSN: education requirements. The first thing you should know is that becoming an RN means passing the NCLEX exam. There is no way around it. To be eligible to sit for the NCLEX you must first earn either an associate degree in nursing (ADN) or a Bachelor of Science in nursing (BSN).The former requires completion of a 21-month program at a career-focused college.he first thing you should know is that becoming an RN means passing the NCLEX exam. There is no way around it. To be eligible to sit for the NCLEX you must first earn either an associate degree in nursing (ADN) or a Bachelor of Science in nursing (BSN). \\
4539659 & That additional level of education means that BSNs generally have more opportunities and larger salaries than RNs. Getting your BSN isn’t just something you do for yourself, though. It also benefits your patients. Studies have shown that BSNs have lower patient mortality rates and lower failure to rescue rates. \\
\multicolumn{2}{c}{\textit{Label 1}} \\[-0.5em]
7696609 & IU Online RN-BSN Mobility Option. The RN to BSN Degree Completion program is designed for the RN to receive a Bachelor of Science degree in nursing (BSN). The program offers a flexible curriculum for the education of professional nurses competent in meeting the current and future health needs of society. \\
\bottomrule
\multicolumn{2}{l}{} \\[-1.5em]

\multicolumn{2}{l}{\textit{Query 1037798: who is robert gray}} \\
\toprule
\multicolumn{2}{c}{\textit{Label 0}} \\[-0.5em]
2360245 & John Gray (philosopher) John Nicholas Gray (born 17 April 1948) is an English political philosopher with interests in analytic philosophy and the history of ideas. He retired in 2008 as School Professor of European Thought at the London School of Economics and Political Science. \\
2360247 & John Gray (I) John Gray was born in Brooklyn, New York, USA. He is a writer, director and producer, known for Reckless (2014) White Irish Drinkers (2010), Ghost Whisperer (2005), The Hunley (1999) Martin and Lewis, and Helter Skelter. He was the youngest recipient ever of an American Film Institute Independant Filmmaker Grant. \\
2868740 & (Redirected from Gary Leroi Gray) Gary LeRoi Gray (born February 12, 1987) is an actor and voice actor involved with movies, television, and animation. He is most recognized for his childhood role as Nelson Tibideaux, the son of Sondra Huxtable Tibideaux and Elvin Tibideaux on the NBC sitcom The Cosby Show. He appeared on the series during its eighth and final season (1991–1992). \\
3914692 & George Gray (born March 11, 1967 in Ballwin, Missouri) is an American Game Show Host, announcer, and comedian. He attended High School in Tucson, Arizona and graduated from the University of Arizona, during which time he played drums with a local band named The Reason Why with two of his friends. \\
3986292 & Robert Michael Bob Parr (born June 25, 1957) is an Emmy Award-winning English-born New Zealander, television personality, private security contractor, counter-terrorism expert and former UK Special Forces operator best known for his role as Team Leader and Executive Producer of Shadow Force on the History Channel. \\
\bottomrule
\multicolumn{2}{l}{} \\[-1.5em]

\multicolumn{2}{l}{\textit{Query 1110199: what is wifi vs bluetooth}} \\
\toprule
\multicolumn{2}{c}{\textit{Label 0}} \\[-0.5em]
291162 & Wi-Fi (or WiFi) is a local area wireless computer networking technology that allows electronic devices to network, mainly using the 2.4 gigahertz (12 cm) UHF and 5 gigahertz (6 cm) SHF ISM radio bands. wireless router allows wired and wireless Ethernet LAN devices to connect to a (usually) single WAN device such as a cable modem or a DSL modem. A wireless router allows all three devices, mainly the access point and router, to be configured through one central utility. \\
3647394 & Electrical \& Electronic Engineer \& Hobbyist, Blogger, Gamer and Tech Enthusiast. If you want to connect your phone/laptop/tablet/pc to a WiFi router, using bluetooth, you can't do that. They might share the same frequency but they can't be connected with each other. \\
421 & Wi-Fi or WiFi is a technology that allows electronic devices to connect to a wireless LAN (WLAN) network, mainly using the 2.4 gigahertz (12 cm) UHF and 5 gigahertz (6 cm) SHF ISM radio bands. \\
68578 & Wi-Fi (or WiFi) is a local area wireless computer networking technology that allows electronic devices to network, mainly using the 2.4 gigahertz (12 cm) UHF and 5 gigahertz (6 cm) SHF ISM radio bands. \\
7842393 & Wi-Fi is a wireless local area network that enables portable computing devices to connect easily to the Internet. Standardized as IEEE 802.11 a,b,g,n, Wi-Fi approaches speeds of some types of wired Ethernet. Wi-Fi has become the de facto standard for access in private homes, within offices, and at public hotspots. \\
\bottomrule
\multicolumn{2}{l}{} \\[-1.5em]

\multicolumn{2}{l}{\textit{Query 19335: anthropological definition of environment}} \\
\toprule
\multicolumn{2}{c}{\textit{Label 3}} \\[-0.5em]
8828697 & Environment. The environment is the world in which the organization operates, and includes conditions that influence the organization such as economic, social-cultural, legal-political, technological, and natural environment conditions. Environments are often described as either stable or dynamic. \\
\multicolumn{2}{c}{\textit{Label 2}} \\[-0.5em]
6512137 & 1 Meaning of environment. Environment means everything around to a living being. Especially the circumstances of life of people or society in their life conditions. It comprises the set of natural, social and cultural values existing in a place and at a particular time, that influence in the life of the human being and in the generations to come. \\
7050803 & The environment is the surroundings of an organism including the physical and chemical environment, and other organisms with which it comes into contact. This term is most frequently used in a human context, often referring to factors affecting our quality of life. Natural Resources. \\
\multicolumn{2}{c}{\textit{Label 0}} \\[-0.5em]
2250973 & Environment. The term environment refers to an organization s natural and human surroundings. An organization s environment extends from within the organization itself to the global system, and includes air, water, land, flora, and fauna (including people), and natural resources of all kinds. \\
6902386 & An environment is the combination of all of the physical, chemical, and biological factors acting upon an organism or an ecological community. The interaction of these factors determines the form and survival of living things and of the environment itself. \\
\bottomrule
\multicolumn{2}{l}{} \\[-1.5em]

\multicolumn{2}{l}{\textit{Query 443396: lps laws definition}} \\
\toprule
\multicolumn{2}{c}{\textit{Label 1}} \\[-0.5em]
4526744 & First: LPS for Protection for buildings and installations against direct strike by lightning. This type of LPS protects the building from damage by direct strike lightning but doesn’t prevent the lightning striking the building. This type of LPS can be divided into:-. Conventional lightning protection system, \\
\multicolumn{2}{c}{\textit{Label 0}} \\[-0.5em]
119472 & Lipopolysaccharide (LPS) is the major component of the outer leaflet of the outer membrane of Gram-negative bacteria. The LPS molecule is composed of two biosynthetic entities: the lipid A-core and the O-polysaccharide (O-antigen). \\
2143845 & A foreign limited partnership is a partnership formed under the laws of any state other than this state or under the laws of a foreign country and having one or more general partners and one or more limited partners.o register cancellation or dissolution of the limited partnership, the company must complete Limited Partnership Certificate of Cancellation (Secretary of State Form LP-4/7) and submit it to the Secretary of State. \\
2143847 & DEFINITION of 'Limited Partnership-LP'. Two or more partners united to conduct a business jointly, and in which one or more of the partners is liable only to the extent of the amount of money that partner has invested.Limited partners do not receive dividends, but enjoy direct access to the flow of income and expenses.EFINITION of 'Limited Partnership-LP'. Two or more partners united to conduct a business jointly, and in which one or more of the partners is liable only to the extent of the amount of money that partner has invested. Limited partners do not receive dividends, but enjoy direct access to the flow of income and expenses. \\
6863145 & The Last Planner System is a production planning system designed to produce predictable work flow and rapid learning in programming, design, construction and commissioning of projects. The Last Planner System (LPS) is developed by Glenn Ballard and Greg Howell as a production planning and control system to assist in smoothing variations in construction work flow, developing planning foresight, and reducing uncertainty in construction operations. The \\
\bottomrule
\multicolumn{2}{l}{} \\[-1.5em]

\multicolumn{2}{l}{\textit{Query 1121709: what are the three percenters?}} \\
\toprule
\multicolumn{2}{c}{\textit{Label 1}} \\[-0.5em]
3423064 & A few things you should consider... Posted on April 13, 2014 by David Robertson in Constitution Issues // 2 Comments. There have been a lot of questions lately in regard to the term “Three Percenter”. The term comes up every once in awhile, especially when perceived injustices are committed by the federal government. \\
3423065 & A few things you should consider... There have been a lot of questions lately in regard to the term “Three Percenter”. The term comes up every once in awhile, especially when perceived injustices are committed by the federal government. \\
\multicolumn{2}{c}{\textit{Label 0}} \\[-0.5em]
3518653 & what is bond rate – the cost of the bond can depend on several factors. The first, of course, is the type of a bond. For a performance bond, the general starting rate is three percent (3\%) of the contract. This amount will go lower as the contract size grows. For other bonds, the price is typically lower. \\
4836900 & The Internal Revenue Service charges interest when you owe taxes and penalties on your federal tax return. The amount of interest the IRS charges on a tax liability changes quarterly. See below for the most current IRS interest rates. Beginning April 1, 2015, the interest rates for the IRS are:1  three (3) percent for overpayments [two (2) percent in the case of a corporation]; 2  three (3) percent for underpayments; 3  five (5) percent for large corporate underpayments; and.eginning April 1, 2015, the interest rates for the IRS are: 1  three (3) percent for overpayments [two (2) percent in the case of a corporation]; 2  three (3) percent for underpayments; \\
6812164 & Booz Allen has easy access to what Ruttenbur calls the three-letter agencies, including the FBI, the CIA and the Department of Defense, giving the company an especially bright outlook for 2016. BAH stock has gained about 12 percent this year. \\
7370525 & If you have three percent inflation, the central bank can push the policy rate down to one percent, which translates to a negative two percent in real terms. A zero nominal rate would be a powerful negative three percent with three percent inflation. But if inflation is zero, the real rate can be no lower than the nominal rate, hence the “zero bound.” We hear chatter that the ECB may be contemplating a negative nominal rate to deal with this problem. \\
\bottomrule
\multicolumn{2}{l}{} \\[-1.5em]

\multicolumn{2}{l}{\textit{Query 489204: right pelvic pain causes}} \\
\toprule
\multicolumn{2}{c}{\textit{Label 3}} \\[-0.5em]
2268652 & Acute (or sharp) pain starts suddenly and often has a single cause. This type of pain may be a warning that something is wrong. Some causes of acute pelvic pain include: An infection of your uterus, fallopian tubes, or ovaries; this condition is called pelvic inflammatory disease (PID). An infection of the urethra, bladder, or kidneys. \\
4279451 & A common example of chronic pelvic pain is dysmenorrhea or menstrual cramps. Other causes of chronic pelvic pain include endometriosis, adenomyosis, and ovulation pain. Sometimes an illness starts with intermittent pelvic pain that becomes constant over time, this is often a signal that the problem has become worse. \\
6327177 & Some of the more common sources of acute pelvic pain, or pain that occurs very suddenly, may include: 1  ectopic pregnancy: a pregnancy that occurs outside the uterus. 2  pelvic inflammatory disease: an infection of the reproductive organs.3  twisted or ruptured ovarian cyst. 4  miscarriage or threatened miscarriage.elvic pain may have multiple causes, including: 1  inflammation or direct irritation of nerves caused by acute or chronic trauma, fibrosis, pressure or intraperitoneal inflammation. 2  muscular contractions or cramps of both smooth and skeletal muscles. 3  psychogenic factors, which can cause or aggravate pain. \\
6665319 & Any of these events can cause adhesions to form, which may later cause chronic pelvic pain. Infection and Inflammation. Bladder, vaginal and yeast infections, and inflammations from chlamydia, pelvic inflammatory disease (PID) or endometriosis may also cause chronic pelvic pain. The body’s healing response to all of these conditions is to create adhesions. \\
8781574 & Female pelvic pain is typically caused by a medical condition involving the reproductive organs, urinary tract, lower gastrointestinal tract, or muscles of the abdominal wall. Some causes are always short-term (acute), and others can become long-lasting (chronic) unless successfully treated. \\
8781575 & 8 Common Causes of Feminine Pelvic Pain. 2. Endometriosis. Approximately 5 million women suffer with endometriosis, a chronic condition in which cells grow and spread outside the uterus and painfully break down when the uterine lining is shed during your monthly menstrual period. The symptoms of endometriosis include painful abdominal cramps as well as pain in the lower back and legs. \\
\multicolumn{2}{c}{\textit{Label 0}} \\[-0.5em]
533017 & Causes, Diagnosis And Treatment. Right side abdominal pain is commonly caused by conditions such as appendicitis, gallstones, kidney stones, constipation, ectopic pregnancy, ovarian cyst troubles, endometriosis, Crohn's disease, ulcerative colitis, trapped wind, kidney infection, pulled muscles, hepatitis, and a number of other rarer diseases. \\
\bottomrule
\multicolumn{2}{l}{} \\[-1.5em]

\multicolumn{2}{l}{\textit{Query 1112341: what is the daily life of thai people}} \\
\toprule
\multicolumn{2}{c}{\textit{Label 2}} \\[-0.5em]
3118343 & Twice a day, at 08:00 and again at 18:00, the national anthem is played by all Thai media outlets. Thais stop what they are doing and stand at attention to pay homage to the flag during the anthem. Students in school stand in front of the raised flag and sing the national anthem at 08:00 every school day. \\
7079776 & In the evening, the Thai style of eating often includes a variety of dishes including a combination of flavors and textures (for example, a stir fried dish, a vegetable dish, a soup and fish) served family-style with each person scooping up a little food to put in their own bowls of steamed rice. \\
\multicolumn{2}{c}{\textit{Label 1}} \\[-0.5em]
3524227 & Thai cuisine is essentially a marriage of centuries-old Eastern and Western influences harmoniously combined into something uniquely Thai. Characteristics of Thai food depend on who cooks it, for whom it is cooked, for what occasion, and where it is cooked. Dishes can be refined and adjusted to suit all palates. Originally, Thai cooking reflected the characteristics of a waterborne lifestyle. Aquatic animals, plant and herbs were major ingredients. \\
8606161 & There are over 29,000 temples in Thailand and the daily routine of the monks in all of them is pretty much the same. 4.00 am-The monks wake up and meditate for one hour, followed by one hour of chanting.6.00 am-The monks walk barefoot around the neighbourhood while the local people make merit by offering them food.his is the last solid food they are allowed to consume until sunrise the following morning. 1.00 pm-Classes in Buddhist teaching begin. Some monks may attend school outside the temple. 6.00 pm-A two-hour session of meditation and prayer begins. 8.00 pm-The monks retire to do homework. \\
\multicolumn{2}{c}{\textit{Label 0}} \\[-0.5em]
3493341 & Food and Economy. Food in Daily Life. Rice is the staple of the daily diet, along with manioc and other root vegetables, plantains, fresh and dried fish, and milk from grated coconuts. Food taboos provide a way to establish connections and acknowledge identity. \\
4461904 & Food in Daily Life. Bread and potatoes are the traditional staple foods. Most meals include, pork, chicken, or beef, and Seafood is popular in the northern part of the country. The national drink is beer, but wine is imported in large quantities. In northern cities, popular dishes include mussels with fries and waterzooi a broth of vegetables and meat or fish. \\
4927748 & What we want are places that serve authentic and delicious Thai food, where the Thais themselves love to eat. We have been in Bangkok several times, often staying for long periods of a few months at a time, and have tried a lot of Thai restaurants both indoors and on the streets. \\
7079779 & However, Thai restaurants in the United States often serve soup and salad first to match Western norms. Noodle dishes are the sandwiches of Thai cuisine. Pad Thai, pad see ew and noodle soups are what you grab for lunch or for a quick solo dinner. In the United States, they're on the dinner menus of Thai restaurants, but that's not their traditional role. If you're looking for a salt shaker, use prik nam pla. \\
8139250 & Daily Life of a Luk Thep Doll in Thailand November 20, 2017 by Sukruti Anah Staneley Petch is Thailand’s first known Luk Thep doll – a plastic doll manufactured in the way any doll might be. The only difference being that the doll has a living soul, believed to change people’s lives; in Mama Ning’s case, it was the spirit of her son. \\
\bottomrule
\multicolumn{2}{l}{} \\[-1.5em]

\multicolumn{2}{l}{\textit{Query 1113437: what is physical description of spruce}} \\
\toprule
\multicolumn{2}{c}{\textit{Label 3}} \\[-0.5em]
4473157 & The white spruce is a large coniferous evergreen tree which grows normally to 15 to 30 metres (49 to 98 ft) tall, but can grow up to 40 m (130 ft) tall with a trunk diameter of up to 1 m (3.3 ft).The bark is thin and scaly, flaking off in small circular plates 5 to 10 centimetres (2.0 to 3.9 in) across.he cones are pendulous, slender, cylindrical, 3 to 7 cm (1.2 to 2.8 in) long and 1.5 cm (0.59 in) wide when closed, opening to 2.5 cm (0.98 in) broad. They have thin, flexible scales 15 mm (0.59 in) long, with a smoothly rounded margin. They are green or reddish, maturing to pale brown 4 to 8 months after pollination. \\
4498867 & Spruces are large trees, from about 20–60 metres (about 60–200 feet) tall when mature, and can be distinguished by their whorled branches and conical form. The needles, or leaves, of spruce trees are attached singly to the branches in a spiral fashion, each needle on a small peg-like structure called a pulvinus.hey are also used by the larvae of gall adelgids (Adelges species). In the mountains of western Sweden scientists have found a Norway spruce tree, nicknamed Old Tjikko, which by reproducing through layering has reached an age of 9,550 years and is claimed to be the world's oldest known living tree. \\
5514586 & The white spruce is a large coniferous evergreen tree which grows normally to 15 to 30 m (50 to 100 ft) tall, but can grow up to 40 m (130 ft) tall with a trunk diameter of up to 1 m (3.3 ft). The bark is thin and scaly, flaking off in small circular plates 5 to 10 cm (2 to 4 in) across. \\
5938420 & 39 Species with 604 Trinomials. Picea a genus of about 35 species of coniferous evergreen trees in the family Pinaceae, found in the northern temperate and boreal regions of the earth.Spruces are large trees, from 20–60 metres (66–200 ft) tall when mature, and can be distinguished by their whorled branches and conical form.or the same reason, some (particularly Picea abies and Picea omorika) are also extensively used as Christmas trees. The word “spruce” entered the English language from Old French Pruce, the name of Prussia. \\
5938425 & Seeds. The white spruce is a large coniferous evergreen tree which grows normally to 15 to 30 metres (49 to 98 ft) tall, but can grow up to 40 m (130 ft) tall with a trunk diameter of up to 1 m (3.3 ft). The bark is thin and scaly, flaking off in small circular plates 5 to 10 centimetres (2.0 to 3.9 in) across.he bark of mature white spruce is scaly or flaky, gray-brown (Brayshaw 1960) or ash-brown (Harlow and Harrar 1950), but silvery when freshly exposed. Resin blisters are normally lacking, but the Porsild spruce Picea glauca var. \\
8128793 & Common Name: Blue Spruce, Colorado Spruce. Picea abies Common Name: Norway Spruce . About the Spruce Tree . Spruce trees are coniferous (needle-leafed) evergreens. There are about 35 species in the genus Picea, native to the northern hemisphere. The name spruce is from the Latin pix, meaning pitch. These trees often live to a very old age. The oldest spruce on record is a Picea engelmannii that reached 852 years and is still thriving. \\
\multicolumn{2}{c}{\textit{Label 2}} \\[-0.5em]
1896943 & Norway spruce is a large, fast-growing evergreen coniferous tree growing 35–55 m (115–180 ft) tall and with a trunk diameter of 1 to 1.5 m (39 to 59 in). It can grow fast when young, up to 1 m (3 ft) per year for the first 25 years under good conditions, but becomes slower once over 20 m (65 ft) tall. \\
742042 & Common Name(s): Spruce Pine. Scientific Name: Pinus glabra. Distribution: Southeastern United States (coastal plain) Tree Size: 65-100 ft (20-30 m) tall, 2-3 ft (.6-1 m) trunk diameter. Average Dried Weight: 33 lbs/ft3 (525 kg/m3) Specific Gravity (Basic, 12\% MC): .42, .52. Janka Hardness: 700 lbf (3,110 N) \\
\multicolumn{2}{c}{\textit{Label 0}} \\[-0.5em]
8503451 & Spruce trees and shrubs are classified in the genus Picea, which includes 35 species. It is considered to be part of the Pinaceae family, which also includes pine trees, fir trees, cedars, hemlocks, larches and a few other species. \\
\bottomrule
\multicolumn{2}{l}{} \\[-1.5em]

\multicolumn{2}{l}{\textit{Query 855410: what is theraderm used for}} \\
\toprule
\multicolumn{2}{c}{\textit{Label 3}} \\[-0.5em]
8651773 & Thera-Derm topical. Uses. This medication is used as a moisturizer to treat or prevent dry, rough, scaly, itchy skin and minor skin irritations (e.g., diaper rash, skin burns from radiation therapy). Emollients are substances that soften and moisturize the skin and decrease itching and flaking. Some products (e.g., zinc oxide, white petrolatum) are used mostly to protect the skin against irritation (e.g., from wetness). Dry skin is caused by a loss of water in the upper layer of the skin. \\
\multicolumn{2}{c}{\textit{Label 0}} \\[-0.5em]
3113274 & For example, if I pointed to my watch and used the wh-q facial expression it would mean that I'm asking you What time is it? You've probably already used this sign many times in your life. WHAT (huh version) In the picture below I'm doing a general gesture for what while using the WHAT facial expression. \\
3618639 & It's used to treat actinic keratosis, as well as acne, rosacea, skin cancer, sun damage, oily skin, wrinkles, warts, psoriasis, and enlarged sebaceous glands. Psoriasis Psoriasis is a long-term skin condition that may cause large plaques of red, raised skin, flakes of dry skin, and skin scales. \\
4492180 & Hereby is used in connection with what is present with the speaker or what the speaker has just mentioned or the speaker's present words (i.e. the present document); thereby is used in connection with what is at a distance from the speaker or what has been mentioned by the speaker or by someone else. \\
4708042 & Tresaderm is one of the most commonly prescribed pet drugs on the market today. It is used to treat ear and skin conditions even when the underlying cause is not known. Only Tresaderm contains the unique combination of active ingredients that makes it the proven solution for dogs and cats, including: 1  Antifungal thiabendazole – controls the most common forms of fungus that cause infection. \\
4708045 & Tresaderm is sometimes prescribed by vets to treat bacterial, inflammatory and/or fungal skin infections in dogs as well as inflammation of the ear. The medicine is sold as a liquid solution which contains dexamethasone neomycin sulfate and thiabendazole. \\
4708046 & Dermatologic Solution TRESADERM is indicated as an aid in the treatment of certain bacterial, mycotic, and inflammatory dermatoses as well as otitis externa in dogs and cats. \\
4708047 & When you should give your pet Tresaderm: Tresaderm is a dermatologic solution indicated as an aid in the treatment of certain bacterial, fungal and inflammatory skin disorder and otitis externa in dogs and cats. How Tresaderm should be used: Clean the affected area prior to use. The amount of medication to apply and the frequency of treatment are dependent on the severity and extent of the lesions. The typical dose for the ear is 5 to 15 drops instilled into the ear twice a day. \\
5550193 & Use the language of coping and the strategies listed in the above categories, which tell us what to do and what not to do. 1  Be aware of the coping strategies you use, and evaluate them honestly. 2  Decide what sort of social role model you are. 3  Expand your coping skills.se the language of coping and the strategies listed in the above categories, which tell us what to do and what not to do. 1  Be aware of the coping strategies you use, and evaluate them honestly. 2  Decide what sort of social role model you are. 3  Expand your coping skills. \\
7066410 & CONTACT US. Want a quick way to see what GPO’s are applied to your local system, just using built in utilities? Using the GUI to manually view what settings are applied is awkward and slow. ?Use the following commands to see what policies are being handed down to the system you’re on and what they’re enforcing. \\
8539176 & To be effective using an interface you've designed, people must be able to recognize what it is, care about why they would use it, understand what the interface is helping them interact with, predict what will happen when they use it, and then successfully interact with it. \\
8651778 & What is Tresaderm: Tresaderm is a dermatologic solution used in dogs and cats as an aid in the treatment of certain acute or chronic bacterial, fungal and inflammatory skin disorders as well as otitis externa. Tresaderm may also be used for purposes other than those listed. \\
\bottomrule
\multicolumn{2}{l}{} \\[-1.5em]

\end{longtable}
\normalsize

\twocolumn

\onecolumn
\newcolumntype{Y}{>{\small\raggedright\arraybackslash}X}
\renewcommand{\arraystretch}{1.5}
\noindent

\small
\begin{longtable}{>{\setlength\hsize{.15\hsize}}X>{\setlength\hsize{1.75\hsize}}X}
\multicolumn{2}{l}{\textit{Query 1051399: who sings monk theme song}} \\
\toprule
\multicolumn{2}{c}{\textit{Label 0}} \\[-0.5em]
2894408 & Thiele then contacted Curtis Stigers to write lyrics and sing the theme, who recorded the song at Cunningham Audio Production in Boise, Idaho. Show creator Kurt Sutter also cowrote the lyrics to the song. When speaking about why he cowrote the theme for the show, Kushner stated that: \\
\bottomrule
\multicolumn{2}{l}{} \\[-1.5em]

\multicolumn{2}{l}{\textit{Query 1108651: what the best way to get clothes white}} \\
\toprule
\multicolumn{2}{c}{\textit{Label 0}} \\[-0.5em]
4304301 & Save. Laundry detergent is what keeps our clothes and linens looking their best. Detergent comes in various formulas to whiten whites, brighten brights, smell fresh or have no smell at all, even be completely hypoallergenic. \\
\bottomrule
\multicolumn{2}{l}{} \\[-1.5em]

\multicolumn{2}{l}{\textit{Query 1122767: what amino produces carnitine}} \\
\toprule
\multicolumn{2}{c}{\textit{Label 0}} \\[-0.5em]
8039587 & Carnitine is a quaternary ammonium compound involved in metabolism in most mammals, plants and some bacteria. Carnitine may exist in two isomers, labeled D-carnitine and L-carnitine, as they are optically active. At room temperature, pure carnitine is a white powder, and a water-soluble zwitterion with low toxicity. Carnitine only exists in animals as the L-enantiomer, and D-carnitine is toxic because it inhibits the activity of L-carnitine. Carnitine was discovered in 1905 as a result of its hi \\
\bottomrule
\multicolumn{2}{l}{} \\[-1.5em]

\multicolumn{2}{l}{\textit{Query 324585: how much money do motivational speakers make}} \\
\toprule
\multicolumn{2}{c}{\textit{Label 0}} \\[-0.5em]
8260792 & According to a 2013 survey (.pdf file) of 94,000 Etsy sellers, 81\% said they opened their shops as a creative outlet, while 68\% cited supplemental income as a motivating factor. “They started because they love it, and then they figured out how to make money from doing what they love,” Katrina adds. \\
\bottomrule
\multicolumn{2}{l}{} \\[-1.5em]

\multicolumn{2}{l}{\textit{Query 332593: how often to button quail lay eggs}} \\
\toprule
\multicolumn{2}{c}{\textit{Label 0}} \\[-0.5em]
8749230 & In the right conditions, each hen can lay 300+ per year. They begin laying eggs in 6 to 8 weeks and the eggs only take 17 days to incubate. Because Coturnix quail mature extremely quickly, the waiting period to enjoy the fruits of your labors are very quick. \\
\bottomrule
\multicolumn{2}{l}{} \\[-1.5em]

\multicolumn{2}{l}{\textit{Query 42255: average salary for dental hygienist in nebraska}} \\
\toprule
\multicolumn{2}{c}{\textit{Label 0}} \\[-0.5em]
5783300 & How much does a dental hygienist make? According to the May 2012 records from the Bureau of Labor Statistics, the mean annual dental hygienist salary is \$70,700, which is equivalent to a mean hourly wage of \$33.99. The bottom ten percent in the occupation earns an annual salary of \$46,540, while the top ten percent make an average \$96,280 each year. Dental hygienists employed in dental offices earn an average \$71,000, while those employed in physician offices make significantly less at a mean salary of \$65,310. Within the United States, the highest paid dental hygienists are employed in California, where they earn an annual salary of \$93,280. Following closely behind, those in Nevada make \$91,350 and dental hygienists in Washington make an annual mean salary of \$90,540. \\
\bottomrule
\multicolumn{2}{l}{} \\[-1.5em]

\multicolumn{2}{l}{\textit{Query 555530: what are best foods to lower cholesterol}} \\
\toprule
\multicolumn{2}{c}{\textit{Label 2}} \\[-0.5em]
6221470 & Top 8 Cholesterol-Lowering Foods. From oats to walnuts, a handful of everyday foods are stepping up to the plate when it comes to battling unhealthy cholesterol. \\
\bottomrule
\multicolumn{2}{l}{} \\[-1.5em]

\multicolumn{2}{l}{\textit{Query 701453: what is a statutory deed}} \\
\toprule
\multicolumn{2}{c}{\textit{Label 1}} \\[-0.5em]
8626142 & With a Statutory Warranty Deed, the seller expressly guarantees the grantor's good, clear title, and often carries the covenants concerning quality such as quiet enjoyment, the right to convey, freedom from encumbrances, and defense of title from all claims. \\
\bottomrule
\multicolumn{2}{l}{} \\[-1.5em]

\multicolumn{2}{l}{\textit{Query 914916: what type of tissue are bronchioles}} \\
\toprule
\multicolumn{2}{c}{\textit{Label 0}} \\[-0.5em]
5061651 & The bronchial tubes divide into smaller air passages (bronchi), and then into bronchioles. The bronchioles end in tiny air sacs called alveoli, where oxygen and carbon dioxide are exchanged. After absorbing oxygen, the blood leaves the lungs and is carried to the heart. Then, it is pumped through your body to provide oxygen to the cells of your tissues and organs. \\
\bottomrule
\multicolumn{2}{l}{} \\[-1.5em]

\multicolumn{2}{l}{\textit{Query 118440: define bmt medical}} \\
\toprule
\multicolumn{2}{c}{\textit{Label 0}} \\[-0.5em]
228375 & Army Basic Training (BMT). Today, the United States Army basic training only takes 9 weeks to transition all you young men and women out there from being a civilian into that of a soldier. \\
3368681 & Basic Military Training, or BMT, is the first step in beginning your Air Force career. There is no separate boot camp for Air Force Reservists. You will train with active-duty and Air National Guard members. Air Force BMT is located at Joint Base San Antonio, part of the former Lackland Air Force Base, in San Antonio, Texas. \\
\bottomrule
\multicolumn{2}{l}{} \\[-1.5em]

\multicolumn{2}{l}{\textit{Query 23849: are naturalization records public information}} \\
\toprule
\multicolumn{2}{c}{\textit{Label 0}} \\[-0.5em]
1449785 & Public records are in fact a wide variety of records that are being kept by the government and are open to the public under the freedom of information act. They include marriage records, divorce records, court records, birth records, death records and many more. In Scotts, MI records are kept mostly on physical files. There is a great rise in the use of electronic copies that are available on the internet. You are advised to look into entries of each record to find out more information on how to obtain them. \\
5925321 & Byron Public Records. Public records are in fact a wide variety of records that are being kept by the government and are open to the public under the freedom of information act. They include marriage records, divorce records, court records, birth records, death records and many more. In Byron, OK records are kept mostly on physical files. There is a great rise in the use of electronic copies that are available on the internet. \\
7431268 & You've probably heard the term public record before. It's a document, or record of information, recorded by a court or governmental agency, that's made available to the public. Some public records, such as tax liens, may appear on your credit report while others, like a marriage license, do not. \\
\bottomrule
\multicolumn{2}{l}{} \\[-1.5em]

\multicolumn{2}{l}{\textit{Query 877809: what metal are hip replacements made of}} \\
\toprule
\multicolumn{2}{c}{\textit{Label 0}} \\[-0.5em]
2213765 & Total hip replacement surgery uses metal, ceramic, or plastic parts to replace the ball at the upper end of the thighbone (femur) and resurface the hip socket in the pelvic bone. Total hip replacement surgery replaces damaged cartilage with new joint material in a step-by-step process.Doctors may attach replacement joints to the bones with or without cement. 1  Cemented joints are attached to the existing bone with cement, which acts as a glue and attaches the artificial joint to the bone.ingle mini-incision total hip replacement for the management of arthritic disease of the hip: A systematic review and meta-analysis of randomized controlled trials. Journal of Bone and Joint Surgery, American Version, 94(20): 1897-1905. \\
3328100 & Total hip replacement surgery uses metal, ceramic, or plastic parts to replace the ball at the upper end of the thighbone (femur) and resurface the hip socket in the pelvic bone. Total hip replacement surgery replaces damaged cartilage with new joint material in a step-by-step process. Doctors may attach replacement joints to the bones with or without cement. Cemented joints are attached to the existing bone with cement, which acts as a glue and attaches the artificial joint to the bone. \\
5850349 & Hip Replacement Surgery. Guide. Total joint replacement involves surgery to replace the ends of both bones in a damaged joint to create new joint surfaces. Total hip replacement surgery uses metal, ceramic, or plastic parts to replace the ball at the upper end of the thighbone (femur) and resurface the hip socket in the pelvic bone. Total hip replacement surgery replaces damaged cartilage with new joint material in a step-by-step process . Doctors may attach replacement joints to the bones with or without cement. Cemented joints are attached to the existing bone with cement, which acts as a glue and attaches the artificial joint to the bone. \\
\bottomrule
\multicolumn{2}{l}{} \\[-1.5em]

\multicolumn{2}{l}{\textit{Query 1121353: what can you do about discrimination in the workplace in oklahoma city}} \\
\toprule
\multicolumn{2}{c}{\textit{Label 1}} \\[-0.5em]
1727596 & 1 Communicate clearly with your employees about what type of behavior is lawful and appropriate in the workplace with regard to all forms of discrimination and harassment, including gender identity, and about what consequences your employees will face if they violate the law. \\
4593222 & Workplace Discrimination Laws and Policies. Under the patchwork of state and local employment law that prohibits employment discrimination based on gender identity and sexual orientation more than three of every five citizens live in jurisdictions that do not provide such protections, and they are needed. \\
\multicolumn{2}{c}{\textit{Label 0}} \\[-0.5em]
5540042 & Getty Images. Discriminating against your employees can cost you. Workplace discrimination against employees based on race, gender or sexual orientation costs businesses an estimated \$64 billion annually, a recent report from the Center For American Progress finds.etty Images. Discriminating against your employees can cost you. Workplace discrimination against employees based on race, gender or sexual orientation costs businesses an estimated \$64 billion annually, a recent report from the Center For American Progress finds. \\
5606063 & The crippling effects of workplace discrimination include poor work culture and a demoralized workforce, debilitating effect on the individual, negative fallout for the society, and reduced profits for the organization.slide 1 of 7.oss of Focus and Productivity. Another negative effect of workplace employee discrimination is loss of focus and wasting of time. Discrimination in the workplace is not just the focus of the victim, but also tends to take priority in various meetings. \\
6381283 & It is important to note that not all forms of discrimination are illegal. There are several federal laws that prohibit discrimination in the workplace, including: 1  Equal Pay Act of 1963: Requires that men and women receive equal pay for doing the same or similar jobs. \\
\bottomrule
\multicolumn{2}{l}{} \\[-1.5em]

\multicolumn{2}{l}{\textit{Query 67316: can fever cause miscarriage early pregnancy}} \\
\toprule
\multicolumn{2}{c}{\textit{Label 0}} \\[-0.5em]
156062 & Miscarriage: Bleeding can be a sign of miscarriage, but does not mean that miscarriage is imminent. Studies show that anywhere from 20-30\% of women experience some degree of bleeding in early pregnancy. Approximately half of pregnant women who bleed do not have miscarriages. \\
4163707 & Miscarriage: Bleeding can be a sign of miscarriage, but does not mean that miscarriage is imminent. Studies show that anywhere from 20-30\% of women experience some degree of bleeding in early pregnancy. Approximately half of pregnant women who bleed do not have miscarriages. Approximately 15-20\% of all pregnancies result in a miscarriage, and the majority occur during the first 12 weeks. Signs of Miscarriage include: \\
6221796 & Bleeding during early pregnancy is sometimes a sign of miscarriage. About half the time, bleeding during pregnancy is a sign of miscarriage, and the other half of the time it is not. So, a good rule of thumb to remember is that if the bleeding is enough to be a cause of concern for you, you should see your doctor. \\
6764673 & Miscarriage: Bleeding can be a sign of miscarriage, but does not mean that miscarriage is imminent. Studies show that anywhere from 20-30\% of women experience some degree of bleeding in early pregnancy.Approximately half of pregnant women who bleed do not have miscarriages.iscarriage: Bleeding can be a sign of miscarriage, but does not mean that miscarriage is imminent. Studies show that anywhere from 20-30\% of women experience some degree of bleeding in early pregnancy. \\
7513151 & First Half of Pregnancy: Miscarriage: Bleeding can be a sign of miscarriage, but does not mean that miscarriage is imminent. Studies show that anywhere from 20-30\% of women experience some degree of bleeding in early pregnancy. Approximately half of pregnant women who bleed do not have miscarriages. Approximately 15-20\% of all pregnancies result in a miscarriage, and the majority occur during the first 12 weeks. \\
\bottomrule
\multicolumn{2}{l}{} \\[-1.5em]

\end{longtable}
\normalsize

\twocolumn

\end{document}